\documentclass[pmlr]{jmlr}

\usepackage{longtable}

\usepackage{booktabs}

\usepackage[load-configurations=version-1]{siunitx} 
 \newcommand{\ts}{\textsuperscript}

\theorembodyfont{\upshape}
\theoremheaderfont{\scshape}
\theorempostheader{:}
\theoremsep{\newline}

\jmlrvolume{1}
\jmlryear{2020}
\jmlrworkshop{NeurIPS2019 Competition \& Demonstration Track}

\title[Sample Efficient Reinforcement Learning in Minecraft]{Sample Efficient Reinforcement Learning through Learning from Demonstrations in Minecraft}

 \author{\Name{Christian Scheller} \Email{christian.scheller@fhnw.ch}\\
    \Name{Yanick Schraner} \Email{yanick.schraner@fhnw.ch}\\
    \Name{Manfred Vogel} \Email{manfred.vogel@fhnw.ch}\\
    \addr Institute for Data Science, University of Applied Sciences Northwestern Switzerland}

\editors{Hugo Jair Escalante and Raia Hadsell}

\begin{document}

\maketitle

\begin{abstract}
  Sample inefficiency of deep reinforcement learning methods is a major obstacle for their use in real-world applications. 
  In this work, we show how human demonstrations can improve final performance of agents on the Minecraft minigame \emph{ObtainDiamond} with only 8M frames of environment interaction.
  We propose a training procedure where policy networks are first trained on human data and later fine-tuned by reinforcement learning. 
  Using a policy exploitation mechanism, experience replay and an additional loss against catastrophic forgetting, our best agent was able to achieve a mean score of 48.
  Our proposed solution placed 3\ts{rd} in the NeurIPS MineRL Competition for Sample-Efficient Reinforcement Learning.
\end{abstract}
\begin{keywords}
Imitation learning, deep reinforcement learning, MineRL competition
\end{keywords}

\section{Introduction}
\label{sec:intro}

The NeurIPS MineRL competition introduced by \citet{guss2019minerlcomp} is focused on the problem of sample efficient reinforcement learning by leveraging human demonstrations. 
The goal of the competition is to solve the \emph{ObtainDiamond} task using 8 million samples from the MineRL environment.
Additionally, agents can learn from a dataset consisting of over 60 million state-action pairs of human demonstrations solving nine distinct complex, hierarchical tasks in the MineRL environment.

Imitation learning is a promising method to address such hard exploration tasks.
In this work, we propose a training pipeline that utilizes human demonstrations to bootstrap reinforcement learning.
Agents are represented as neural networks that predict next actions (policy) and evaluate environment states (value function).
In a first stage, policy networks are trained in a supervised setting to predict recorded human actions given corresponding observations.
These policy networks are then refined in a second stage by reinforcement learning in the MineRL environment.

We show that naive reinforcement learning applied to the supervised trained policies did not lead to improvement within 8M frames.
In contrast, we observe collapsing performance.
We address this problem with four major enhancements:
(1) To improve sample efficiency we make extensive use of experience replay \citep{Lin1992}. 
(2) We prevent catastrophic forgetting and stabilize learning by using CLEAR \citep{NIPS2019_8327}.
(3) We investigate a new mechanism named advantage clipping that allows agents to better exploit good behaviour learned from demonstrations.
(4) We demonstrate that our agents benefit from a separate critic network compared to a combined policy-value network commonly used.
We discuss the major components of our solution in the following sections.

\subsection{Related Work}

The application of imitation learning itself or in combination with reinforcement learning to simplify the exploration problem and improve final performance has been investigated in various challenging domains.
For AlphaGO \citep{Silver2016}, the first reinforcement learning agent to beat the human world champion in the board game Go, Silver et al. applied supervised learning from human demonstration to learn policy- and value-networks that are later refined by reinforcement learning. AlphaStar \citep{Vinyals2019}, the first StarCraft II AI to reach grand master level performance, was initially trained on human demonstrations and later improved in a league, competing with different agents and constantly learning using reinforcement learning.
In the same work Vinyals et al. introduced an upgoing policy gradient that is closely related to self-imitation learning  \citep{pmlr-v80-oh18b} and similar to our advantage clipping.
Both methods restrict updates to only better-than-average trajectories. 
In their work on Deep Q-learning from Demonstrations, \citet{AAAI1816976} successfully incorporated demonstration data in the reinforcement learning loop to improve performance in 11 out of 42 games of the Arcade Learning Environment.
On the same domain, \citet{cruz2017pre} showed that pre-training on demonstrations leads to a reduced training time of reinforcement learning algorithms.
\citet{gao2018reinforcement} introduced a hybrid imitation and reinforcement learning algorithm that learns from imperfect demonstrations to improve performance on tasks in realistic 3D simulations.
%


Experience replay \citep{Lin1992} is a common technique to improve sample efficiency and reduce sample correlation of deep Q-learning algorithms \citep{Mnih2015,Schaul2016,hessel2018rainbow}.
\citet{DBLP:conf/iclr/0001BHMMKF17} showed that experience replay can significantly improve sample efficiency of their actor-critic deep reinforcement learning agents.
\citet{pmlr-v80-espeholt18a} achieved improved performance on tasks in visually complex environments by combining a novel off-policy actor-critic algorithm with experience replay.


With CLEAR, \citet{NIPS2019_8327} showed that experience replay can effectively counter catastrophic forgetting in continual learning.
In contrast, previous research on mitigating forgetting mainly focused on synaptic consolidation approaches \citep{rusu2016progressive,kirkpatrick2017overcoming,schwarz2018progress}.


\subsection{ObtainDiamond}
\label{sec:obtainDia}
\emph{ObtainDiamond} is a Minecraft mini-game with the goal of collecting one piece of diamond within 15 minutes of play time.
The player starts at a random location on a randomly generated map,  without any items.
To mine a diamond, a player has to first craft an iron pickaxe, which itself requires a list of prerequisite items that hierarchically depend on each other.
The player receives a reward for each of these items, whereas subsequent items yield exponentially higher rewards.
The game ends when the player obtains a diamond, dies or reaches the maximum step count of 18000 frames.


The challenges for reinforcement learning agents to succeed in this mini-game are manifold.
The rarity of diamonds (2-10 times less frequent than other ores), the dependence on prerequisite items and the sparsity of the reward signal make naive exploration methods practically infeasible.
Agents have to solve long-horizon credit assignment problems, where rewards have to be transported over many time steps.
Besides information about current inventory and equipment, agents perceive the environment through visually complex point-of-view observations, from which an optimal next action must be inferred.
Since maps are generated randomly, agents have to generalize across a virtually infinite number of maps.

The competition introduces further complexity to the problem. 
8 million samples is significantly fewer than what reinforcement learning algorithms traditionally need to master similar problems.
For example, the current state of the art for the Atari-57 benchmark uses 20 billion frames \citep{schrittwieser2019mastering}.
The limited number of frames forces competition entries to be particularly sample efficient.
The hardware- and time-limitation restrict model complexity and algorithms computing power demands.
Although \emph{ObtainDiamond} is a difficult problem for reinforcement learning agents, it is a rather easy for humans. 
Experienced players solve it in less than 15 minutes \citep{ijcai2019-339}.

\section{Methods}

Formally, we consider the \emph{ObtainDiamond} task as a partially observable Markov decision process. 
In order to deal with uncertainty about the current state, we employ long short-term memories (LSTMs) \citep{Hochreiter1997}.
This allows us to reduce the problem to a standard Markov decision process $(\mathcal{S}, \mathcal{A}, \mathcal{P}, \mathcal{R})$, where state $s_t \in \mathcal{S}$ at time step $t$ is given by observation $o_t$ and the current state of the LSTM $h_t$.
Hereby $\mathcal{S}$ is the set of states, $\mathcal{A}$ is the set of actions, $\mathcal{P}(s' \mid s, a)$ is a state transition probability function and $\mathcal{R}(s, a)$ is the reward function. 
The goal is to learn a policy $\pi_\theta(a \mid s)$ that maximizes the expected sum of rewards $\mathbb{E}_{\pi_\theta}[\sum_{t=1}^{T}r_t]$, where $\theta \in \mathbb{R}^n$ is a parameter vector and $T$ is the episode length.



\subsection{Network architecture}
\label{sec:networkArchitecture}

The network architecture for policy and value function is based on the residual model introduced by \citet{pmlr-v80-espeholt18a}.
This architecture has been shown to be effective in visually complex environments such as the DeepMind DMLab-30 environment \citep{pmlr-v80-espeholt18a}, the 
obstacle tower challenge \citep{nichol_2019} and the CoinRun 
environment \citep{DBLP:conf/icml/CobbeKHKS19}.

In this model, spatial inputs are passed to a convolutional neural network with 6 residual blocks, each consisting of 3 convolutional layers to produce a spatial representation.
Non-spatial inputs are concatenated with the agents previously taken action and processed by two dense layers (256 and 64 units respectively) to form a non-spatial representation.
The spatial and non-spatial representations are then concatenated and fed into an LSTM cell with a hidden size of 256.
Since MineRL has a composed action space, we represent each action with an independent policy on top of the LSTM output.

In \sectionref{sec:results:im} we evaluate \emph{craft} and \emph{smelt} policies that use the inventory as additional input, processed by a separate dense two-layer network (256 and 64 units respectively).
The idea behind this modification is that the availability of \emph{craft} and \emph{smelt} actions directly depends on the current inventory.

\subsection{Imitation learning}
\label{sec:imitationLearning}

As a first step, we train policies $\pi_\theta(a | s)$ to predict human actions $a$, given state $s$ on human demonstrations from the MineRL dataset \cite{ijcai2019-339}.
The episode length of the demonstrations is up to 60'000 frames.
To train LSTMs efficiently an episode length reduction becomes necessary.
We use the following subsampling strategy: 

\begin{itemize}
    \item State-action pairs with no-op actions are skipped without compensation.
    \item State-action pairs containing actions that we do not consider necessary for the task, like sneak or sprint, are skipped without compensation.
    \item Consecutive state-action pairs that contain the same action are skipped. 
    Instead we add a \emph{step multiplier} that accounts for the skipped number of frames. 
    This \emph{step multiplier} must be learned by the agent. 
    \item Camera rotations are accumulated and skipped until a threshold of 30 degrees is reached, the rotation direction changes or a new action is issued.
\end{itemize}

Sequences are truncated when a length of 2'000 frames has been reached. 
We only use demonstrations on the tasks \emph{ObtainDiamond}, \emph{ObtainIronPickaxe} and \emph{TreeChop} for training.
We did not consider demonstrations on other tasks to be suitable for imitation learning on the \emph{ObtainDiamond} task.

During training we uniformly sample batches of episodes from the resulting dataset $D$ in the form of sequences of state-action pairs $(s, a) \in D$.
These batches are used to update the parameters $\theta$ by stochastic gradient descent on the cross-entropy loss.


We do not learn a value function estimate from the demonstrations, since the demonstrators policy is very different compared to the policy we obtain from supervised learning.


\subsection{Reinforcement learning}
\label{sec:reinforcementLearning}

We employ the Importance Weighted Actor-Learner Architecture (IMPALA) by \citet{pmlr-v80-espeholt18a} to improve policy $\pi_{\theta} (a \mid s)$ obtained by supervised learning and to approximate the value function with $V_\phi (s)$.
The choice of this architecture is mainly motivated by the following two properties.
First, IMPALA is an off-policy actor-critic method, which enables the use of experience replay. 
Second, as shown by \citet{pmlr-v80-espeholt18a}, asynchronous learners and actors can significantly improve the training throughput.
Slow environments like \emph{ObtainDiamond} with a high variance of update time and slow episode restarts benefit from this asynchrony.  

In \sectionref{sec:results} we show that, within the limited number of frames available, IMPALA applied naively to pre-trained policies exhibits collapsing performance.
In the following sections, we describe our proposed enhancements to prevent performance decline and improve final performance. 

\subsubsection{Separate networks for actor and critic}

In actor-critic algorithms, neural networks for policy and value functions are often represented by individual heads on top of a single neural network. 
This enables parameter sharing but introduces the problem of combined policy and value loss gradients.
With separate networks for actor and critic however, all weights of a network are allocated to either the task of policy or value function approximation.
As shown in \sectionref{sec:results}, we were able to achieve better results by using separate networks for the actor and the critic.

\subsubsection{Experience replay}
\label{sec:experienceReplay}


Similar to \citet{DBLP:conf/iclr/0001BHMMKF17}, we extensively use experience replay to increase sample efficiency of the reinforcement learning training.
The use of experience replay further reduces the correlation between samples.
The hardware restrictions of the competition limit the parallelism to five instances of \emph{ObtainDiamond}, which leads to a low diversity and high correlation of the training data.

In line with \citet{pmlr-v80-espeholt18a}, we employ a ring buffer from which samples are drawn uniformly at random.
In our experiments in \sectionref{sec:results} we evaluated different \emph{replay ratios}, defined as the proportion of the batch size that is sampled from the replay buffer (e.g. a \emph{replay ratio} of 3 corresponds to 3 replay samples per online sample).

\subsubsection{Advantage clipping}
\label{sec:advantageClipping}


We found that policies obtained from imitation learning yield returns with high variance.
This complicates the value function approximation. 
As a result, there is a risk of erroneous value estimates wrongly discouraging desired behaviour.
The idea of advantage clipping is to prevent such destructive updates and to only reinforce  better-than-expected trajectories.
To this end, we introduce a simple mechanism to the policy gradient loss where we clip negative advantages to zero:

\[
-\rho_{t} (\nabla_{\theta} \log \pi_{\theta}(a_t \mid s_t)) \max (r_t + \gamma v_{t + 1} - V_{\phi}(s_t), 0)
\]\\
where $v_t$ is the V-trace target and $\rho_t = \min (\bar{\rho}, \frac{\pi(a_t \mid s_t)}{\mu(a_t \mid s_t)})$ is the truncated importance sampling weight with truncation level $\bar{\rho} = 1$ and behaviour policy $\mu$. 

Clipping the advantage to strictly positive values prevents the policy gradients from reducing probabilities of sampled actions.
As this mechanism suppresses learning from undesired experiences, we consider it primarily useful to stabilize training in high-variance environments.
We believe that advantage clipping could also be scheduled over time.
The choice of clipping threshold, optimal scheduling and theoretical aspects are left for future research.

Advantage clipping is strongly related to self-imitation learning \cite{pmlr-v80-oh18b} which exploits past beneficial decisions and proved its usefulness in difficult exploration tasks.

\subsubsection{Preventing catastrophic forgetting of rarely encountered sub-tasks}
\label{sec:clear}


\figureref{fig:compareRewardDistributionILRL} shows how fine-tuning through reinforcement learning means that agents solve early sub-tasks more frequently but complete later sub-tasks significantly less often.
This reduces the overall performance of the agents, since the reward increases exponentially for later sub-tasks.
In this section we focus on how we can prevent agents from forgetting to solve these highly rewarding tasks. 

This problem is an example of catastrophic forgetting in continual learning.
Agents initialized with supervised trained policies encounter later sub-tasks less frequently than earlier ones.
As a result, most policy updates concern early sub-tasks and override behaviour obtained from demonstrations of later sub-tasks.

We employ CLEAR by \citet{NIPS2019_8327} to prevent such catastrophic forgetting and increase stability of the learning.
CLEAR is a simple but effective method that builds upon the concept of experience replay to reduce forgetting.
It introduces two new components to the IMPALA loss function:
(1) the KL divergence between current and past policy distributions
and (2) the $\ell^2$-norm of the difference between current and past value functions, where past policy distributions and value functions are sampled from an experience replay buffer.

\section{Experiments}
\label{sec:results}

\begin{figure}[t]
    \centering
    \includegraphics[width=1.0\textwidth]{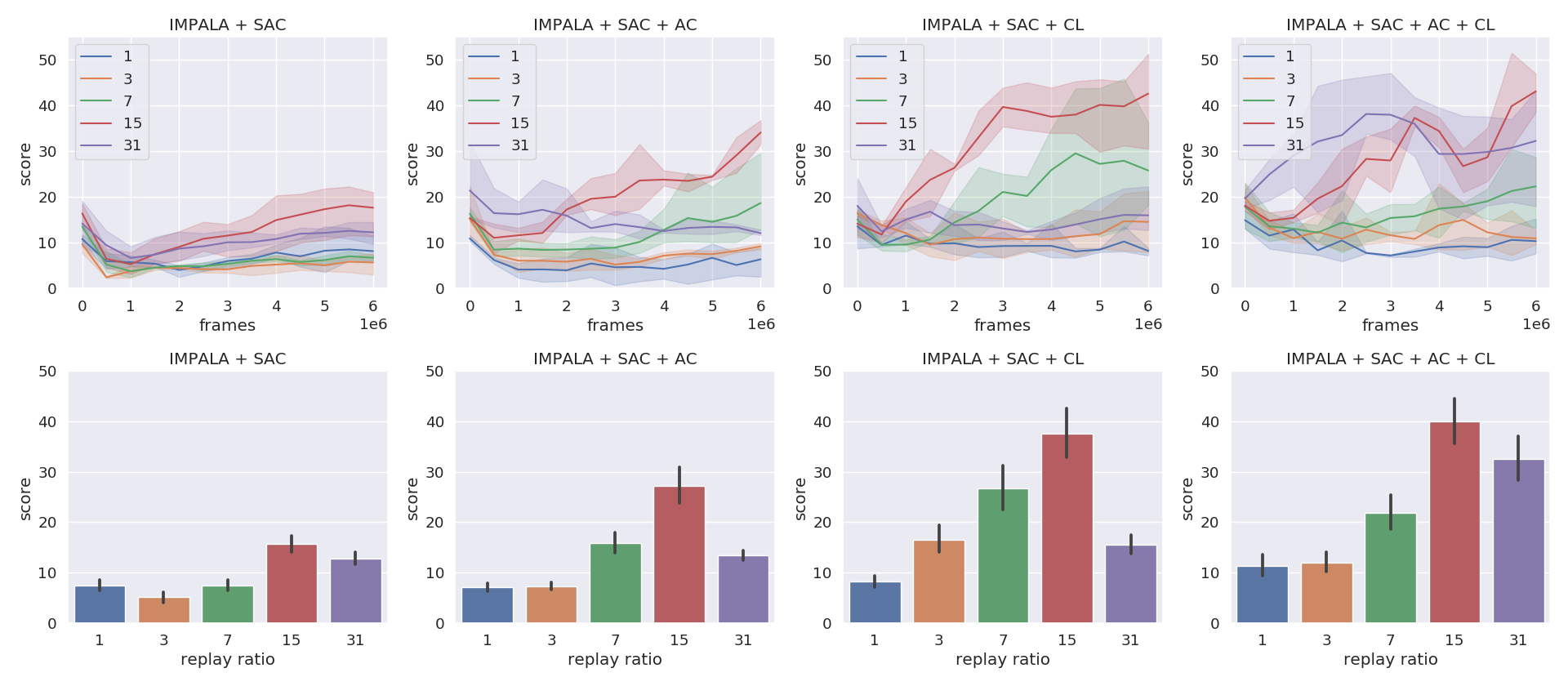}
    \caption{Training performance with replay ratios of 1, 3, 7, 15 and 31. Lines and bars represent the mean score of 3 runs with different random seeds. Bands and error bars correspond to 95\% confidence intervals.}
    \label{fig:replay_buffer_ratio}
\end{figure}

We evaluated the main parts of our solution and how our proposed modifications improve the agent's performance.
In accordance with the MineRL competition rules our experiments complete in less than four days on hardware no more powerful than 6 CPU cores, 56 GiB RAM and a single Nvidia K80 GPU. 
The action space is transformed as follows: camera rotations are discretized to $(-30^\circ, 0^\circ, +30^\circ)$. 
We ran three experiments with different random seeds for each method and evaluated them with 100 episodes. 
We report mean scores, standard deviations, best scores and max episode scores in \tableref{resultTable}.


\subsection{Imitation learning}
\label{sec:results:im}

We ran supervised experiments with the method described in \sectionref{sec:imitationLearning}. 
Each experiment was trained for 125 epochs, with a learning rate of 0.001 and batch size of 16. 
As shown in \tableref{resultTable}, agents benefit from an added inventory input for \emph{craft} and \emph{smelt} policies (CP). 


\begin{figure}[t]
\floatconts
  {fig:compareRewardDistributionILRL}
  {\caption{
  \figureref{fig:rewardFreq} and \figureref{fig:relRewardFreq} compare reward frequencies after (1) supervised learning, (2) fine-tuning with IMPALA, (3) fine-tuning with IMPALA and CLEAR and (4) fine-tuning with IMPALA and advantage clipping. 
  RL fine-tuning improved the performance on early sub-tasks, whereas CLEAR and Advantage Clipping also enabled agents to obtain later rewards more often. 
  These later rewards have an exponentially higher effect on the score, as shown in \tableref{tab:envRewards}.
  }}
  {%
    \subfigure[Reward frequencies]{
        \label{fig:rewardFreq}
        \includegraphics[width=0.35\linewidth]{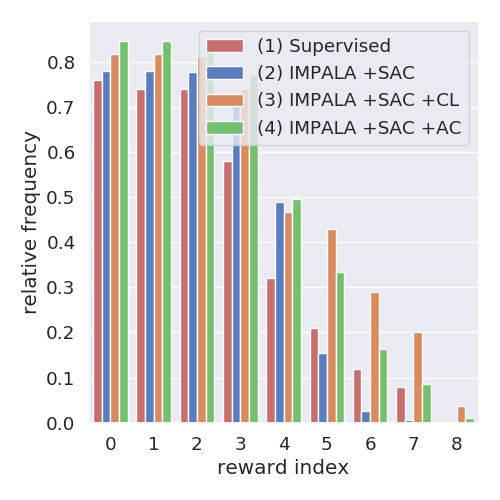}
    }
    \subfigure[Reward frequencies relative to supervised trained policy]{
        \label{fig:relRewardFreq}
        \includegraphics[width=0.35\linewidth]{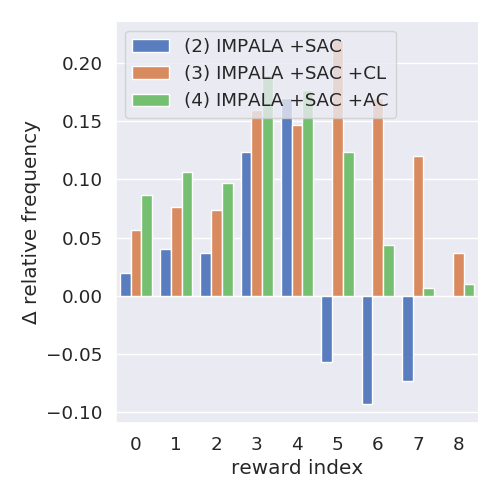}
    }
    \subfigure[Rewards]{
        \label{tab:envRewards}
        \footnotesize
        \begin{tabular}[b]{lc}
            \toprule
            \textbf{Index} & \textbf{Reward} \\
            \midrule
            0 & 1 \\
            1 & 2 \\
            2 & 4 \\
            3 & 4 \\
            4 & 8 \\
            5 & 16 \\
            6 & 32 \\
            7 & 64 \\
            8 & 128 \\
            \bottomrule
            \\ \\
        \end{tabular}
    }
  }
\end{figure}

\subsection{Reinforcement learning}

Reinforcement learning experiments are trained on the maximum allowed number of frames with a batch size of 64.
The agents are initialized with the policies obtained from imitation learning.
In the following paragraphs, we analyze how our proposed enhancements lead to better results.

\paragraph{Experience replay}
We tested experience replay (ER) with replay ratios of 1, 3, 7, 15 and 31.
The results are shown in \figureref{fig:replay_buffer_ratio}.
We observe significantly increased performance with larger replay ratios. A replay ratio of 15 performed best overall.

\paragraph{Separate networks for actor and critic}
We evaluated the separation of actor and critic (SAC) into individual networks.
Both networks use the same architecture as described in \sectionref{sec:networkArchitecture}.
For the first 500'000 frames, we only trained the value network to let it catch up with the policy network.
Our experiments show that separate actor-critics are less prone to performance collapse, but fall short of
the maximum scores that the pre-trained policy achieves.

\paragraph{Advantage clipping}
With advantage clipping (AC) we observed a significant improvement of the mean score.
We find that advantage clipping encourages exploitation of good behaviour of the pre-trained policy and counteracts catastrophic forgetting which is evident in the unchanged maximum score, as shown in \tableref{resultTable}.

\paragraph{CLEAR}
We applied the CLEAR method (CL) with policy-cloning and value-cloning weights of $0.01$ and $0.005$ respectively as proposed by \citet{NIPS2019_8327}.
Like advantage clipping, CLEAR similarly prevents catastrophic forgetting and stabilizes training, but achieves better performance on average. The combination of both methods yields the best results. 
To illustrate the effects of advantage clipping and CLEAR on catastrophic forgetting, we break down the agent's ability to achieve individual rewards in \figureref{fig:compareRewardDistributionILRL}.



\begin{table}
  \caption{
      Ablation study of our proposed methods
      (CP: craft policy, ER: experience replay, SAC: separate actor and critic networks, AC: advantage clipping, CL: CLEAR).
  }
  \label{resultTable}
  \small
  \centering
  \begin{tabular}{lcccc}
    \toprule
    \textbf{Experiment} & \textbf{Mean} & \textbf{Best} & 
    \textbf{Max} \\
    \midrule
    Supervised & 10.3 $\pm$ 4.9 & 13.8 & 131 \\
    +CP & 16.2 $\pm$ 4.1 & 20.2 & 163 \\
    \midrule  
    IMPALA & 2.9 $\pm$ 2.0 & 4.8 & 35 \\
    +ER & 9.6 $\pm$ 0.7 & 10.5 & 67 \\
    +ER +SAC & 15.7 $\pm$ 3.2 & 20.0 & 99 \\
    +ER +SAC +AC & 27.1 $\pm$ 4.6 & 33.6 & 163 \\
    +ER +SAC +CL & 37.5 $\pm$ 4.3 & 41.8 & 163 \\
    +ER +SAC +AC +CL & \textbf{39.9 $\pm$ 8.1} & 47.9 & 163 \\
    \midrule
    Human demonstrations & 1004.0 & - & 1571  \\ 
    \bottomrule
  \end{tabular}
\end{table}


\section{Conclusion}

We introduced a training pipeline that combines imitation learning with reinforcement learning to train agents on \emph{ObtainDiamond}.
Our results 
reveal that performance on highly rewarding later sub-tasks decreased when we applied IMPALA naively to imitation learned policies.
We found that experience replay was crucial to improve the agent's performance when limited to 8M environment frames.
Advantage clipping successfully stabilized the learning and lead to substantially improved policies.
By applying CLEAR, we were able to prevent catastrophic forgetting of rare but highly rewarding behaviour.
We showed that the combination of imitation learning, IMPALA, experience replay with large replay ratios, separate networks for policy and value functions, advantage clipping and CLEAR allowed our agents to achieve a mean score of \textbf{40}. For our best individual agent we observed a mean score of \textbf{48}.

\acks{
    We would like to thank Stephanie Milani and Simon Felix for providing valuable feedback on the previous versions of this manuscript.
}

\newpage

\bibliography{refs}

\end{document}